\pdfoutput=1
\documentclass[5p,twocolumn]{elsarticle}

\usepackage{amssymb}
\usepackage{algorithm}
\usepackage{algpseudocode}

\usepackage{lineno}

\usepackage{bbold}
\newsavebox\CBox
\def\textBF#1{\sbox\CBox{#1}\resizebox{\wd\CBox}{\ht\CBox}{\textbf{#1}}}

\begin{document}
	
\begin{frontmatter}

\title{2D-3D Pose Consistency-based Conditional Random Fields\\ for 3D Human Pose Estimation}

\author[jychang]{Ju~Yong~Chang}
\ead{jychang@kw.ac.kr}

\author[kmlee]{Kyoung~Mu~Lee\corref{cor1}}
\ead{kyoungmu@snu.ac.kr}

\cortext[cor1]{Corresponding author}

\address[jychang]{Department of Electronics and Communications Engineering, Kwangwoon University,\\ 20 Kwangwoon-ro, Nowon-gu, Seoul 01897, KOREA}
\address[kmlee]{Department of Electrical and Computer Engineering, Automation and Systems Research Institute,\\ Seoul National University, 1 Gwanak-ro, Gwanak-gu, Seoul 08826, KOREA}

\begin{abstract}
This study considers the 3D human pose estimation problem in a single RGB image by proposing a conditional random field (CRF) model over 2D poses, in which the 3D pose is obtained as a byproduct of the inference process.
The unary term of the proposed CRF model is defined based on a powerful heat-map regression network, which has been proposed for 2D human pose estimation.
This study also presents a regression network for lifting the 2D pose to 3D pose and proposes the prior term based on the consistency between the estimated 3D pose and the 2D pose.
To obtain the approximate solution of the proposed CRF model, the N-best strategy is adopted.
The proposed inference algorithm can be viewed as sequential processes of bottom-up generation of 2D and 3D pose proposals from the input 2D image based on deep networks and top-down verification of such proposals by checking their consistencies.
To evaluate the proposed method, we use two large-scale datasets: Human3.6M and HumanEva.
Experimental results show that the proposed method achieves the state-of-the-art 3D human pose estimation performance.
\end{abstract}


\end{frontmatter}


\section{Introduction}
\label{Sec:Introduction}

Human pose estimation is one of the most actively investigated problems in computer vision.
Its goal is to infer the configuration of the human body from images or videos.
Recently, single-image 2D human pose estimation has considerably advanced as a result of publicly available benchmark datasets~\citep{Sapp2013,Andriluka2014,Johnson2010} and discriminative methods such as deformable part models~\citep{Felzenszwalb2008,Andriluka2009,Yang2011,Dantone2013,Pishchulin2013} and convolutional neural networks (CNNs)~\citep{Toshev2014,Chen2014a,Tompson2014,Tompson2015,Carreira2016,Yang2016,Wei2016,Newell2016}.
However, 3D human pose estimation from single images remains extremely challenging due to inherent ambiguities~\citep{Lee1985} in recovering 3D information from a 2D image.
Other difficulties include large appearance variations, various types of body shape, (self-)occlusions, and huge solution space.
Recent single-image 3D human pose estimation approaches can be broadly classified into two categories: prediction-based approaches and optimization-based approaches.
The \emph{prediction-based approaches}~\citep{Ionescu2014,Ionescu2014a,Li2014,Tekin2016,Brau2016} exploit training data to find a regression function that can directly generate a 3D pose from an input 2D image.
The \emph{optimization-based approaches}~\citep{Simo-Serra2012,Simo-Serra2013,Kostrikov2014,Li2015,Yasin2016,Zhou2016,Bogo2016} attempt to minimize an energy function including the prior terms that are usually based on 3D pose statistics.

In this study, we propose a new 3D human pose estimation method based on a conditional random field (CRF) framework with a \emph{high-order 2D-3D pose consistency prior}.
Our CRF defines the probability distribution over 2D human poses rather than 3D poses.
The unary likelihood term is defined by using the 2D joint heat maps that are produced by conventional CNN-based 2D human pose estimation approaches~\citep{Tompson2014,Tompson2015,Pfister2015,Pishchulin2016,Insafutdinov2016}.
The high-order 2D-3D pose consistency term is defined by the following steps.
First, we directly estimate 3D pose from 2D pose using the 2D-to-3D pose-lifting network that can be obtained by training with ground-truth 2D and 3D pose data.
Second, we re-project the estimated 3D pose onto the 2D image and then compare the re-projected 2D pose with the original 2D pose to measure the consistency by computing their differences.
If the input 2D pose is normal and probable, then this consistency should be high.
If otherwise, the consistency should be low.
By inferring the maximum a posteriori (MAP) estimate of the proposed CRF, we can find the most probable 2D pose and its corresponding 3D pose as a byproduct.

Therefore, the 2D-to-3D pose-lifting network plays a key role in the proposed method.
Previous methods for 2D-to-3D pose lifting~\citep{Ramakrishna2012,Wang2014,Akhter2015,Zhou2015} are usually based on time-consuming 3D reconstruction processes and assume orthographic camera projection, which results in suboptimal performance.
Therefore, we propose the use of a multilayer perceptron (MLP) that is a simple feedforward neural network.
This network directly regresses the 3D pose from the input 2D pose with high efficiency and can produce more accurate estimates by considering perspective projection.

\begin{figure}[t]
\centering
\includegraphics[width=1.0\linewidth]{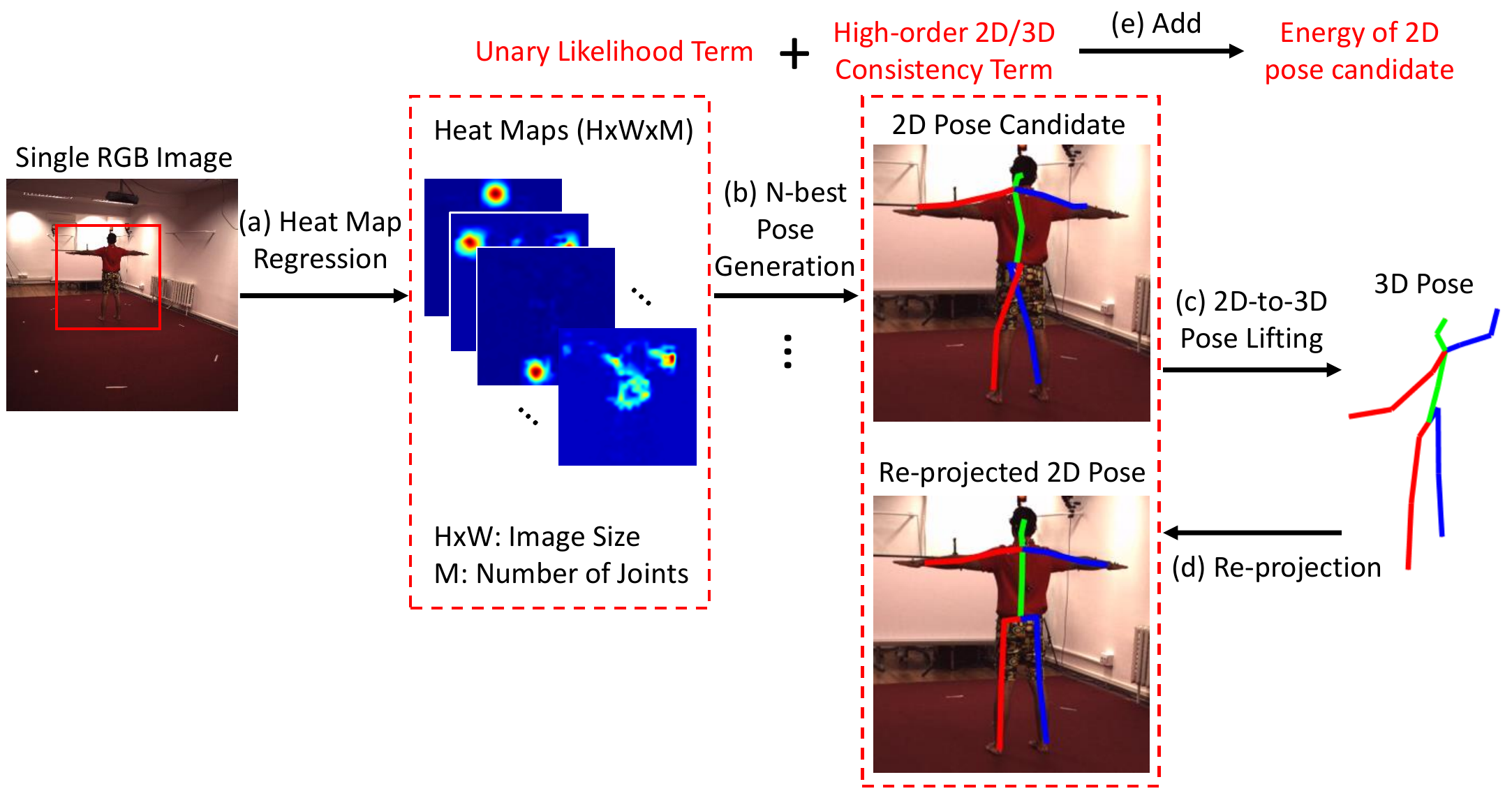}
\caption{Overview of the proposed method.}
\label{Fig:Overview}
\end{figure}

Our CRF model requires the CNN-based heat map regression and MLP-based 2D-to-3D pose-lifting networks.
Both are independently trained using the ground-truth RGB image, 2D pose, and 3D pose data.
Inferring the exact MAP estimate of the proposed CRF model is a highly difficult task because of high-order prior term.
Thus, we adopt the N-best strategy~\citep{Park2011,Cherian2014} to find the approximate solution.
An overview of the proposed method is illustrated in Figure~\ref{Fig:Overview}.
From an input RGB image, per-joint heat maps are generated using the heat map regression network, as shown in Figure~\ref{Fig:Overview}(a).
The heat maps serve as the unary term describing the likelihood of each joint occurring in the 2D spatial location.
Then, 2D pose candidates are obtained by applying the N-best pose generating procedure to the heat maps, as shown in Figure~\ref{Fig:Overview}(b).
For each 2D pose candidate, we use the 2D-to-3D pose-lifting network to produce the 3D pose estimate, which is then re-projected onto the 2D image space as shown in Figure~\ref{Fig:Overview}(c) and (d).
The re-projected 2D pose is compared to the original 2D pose candidate, which results in a high-order 2D-3D consistency term.
By adding the unary likelihood term and 2D-3D consistency term, we can obtain the energy of the 2D pose candidate, as shown in Figure~\ref{Fig:Overview}(e).
Finally, we find the minimum energy among the 2D pose candidates to obtain the optimal 2D pose and its corresponding 3D pose.

The main contributions of this work are as follows:
\begin{itemize}
\item In this study, we propose a new CRF model with a novel 2D pose prior term.
Unlike the conventional priors that explicitly model the probability distribution of the 2D pose, our CRF model implicitly measures the plausibility of the 2D pose by computing the point estimate of the 3D pose and the consistency between the 2D and 3D poses.
Our new 2D-3D pose consistency-based CRF can be combined with the N-best strategy to obtain the approximate solution with high efficiency because the optimization process relies only on two feedforward networks and simple arithmetic operations.

\item We propose a simple but powerful 2D-to-3D pose-lifting method based on the MLP, which has two roles.
First, it is used to constructively define our 2D-3D pose consistency prior.
Second, by computing the prior of a 2D pose, its corresponding 3D pose can be automatically obtained as a byproduct through the proposed pose-lifting network.
The proposed network does not require the assumption of orthographic projection and involved optimization processes but achieves state-of-the-art 2D-to-3D pose-lifting performance.

\item We have conducted thorough experiments on two real datasets: Human3.6M~\citep{Ionescu2014} and HumanEva~\citep{Sigal2010}.
We compare the proposed approach with recent 3D human pose estimation methods and show that ours produces the state-of-the-art results.
\end{itemize}

The remainder of this paper is organized as follows.
We review the related works in Section~2.
The proposed CRF model and inference procedure are presented in Sections~3 and~4, respectively.
We provide the experimental results in Section~5 and the concluding remarks in Section~6.

\section{Related Works}
\label{Sec:RelatedWorks}

\subsection{3D Human Pose Estimation from 2D Pose}

A group of methods have tried to recover 3D human poses from 2D image landmarks.
From the 2D images, the landmarks are usually given by manual annotation or automatic extraction using 2D human pose estimation methods.
All recent 2D-to-3D pose-lifting methods~\citep{Ramakrishna2012,Wang2014,Akhter2015,Zhou2015} use the 3D shape prior enforcing that valid 3D human shape variations should be represented by a linear combination of basis vectors.
The 3D shape and viewpoint (i.e., camera extrinsic parameters) are then obtained by adopting a 3D-to-2D shape fitting process in which 2D re-projection errors are minimized.
In~\citep{Ramakrishna2012}, a greedy orthogonal matching pursuit algorithm is proposed to reconstruct the shape and viewpoint from manually labeled 2D joints while encouraging anthropometric regularity.
In~\citep{Wang2014}, an alternating direction method, which alternately updates the 3D shape and camera parameters, is presented and applied to the inaccurate 2D joints detected by a 2D pose estimator.
In~\citep{Akhter2015}, a physically motivated prior based on pose-dependent joint angle limits is learned from a new dataset that includes an extensive variety of stretching poses.
In~\citep{Zhou2015}, the authors focus on nonconvexity in joint optimization of 3D shape and viewpoint, and propose a convex relaxation approach in which the joint estimation problem is formulated as a convex program and an efficient algorithm is developed on the basis of the alternating direction method of multipliers.
All these methods assume orthographic camera projection and obtain the 3D human pose by estimating the 3D shape and viewpoint separately.
Our method does not require the orthographic assumption and directly generates the 3D human pose in camera coordinates.

\subsection{3D Human Pose Estimation from Single Image}

Numerous studies have focused on 3D human pose estimation from single images.
Early approaches perform automatic discriminative prediction of 3D pose from various image features~\citep{Rosales2001,Agarwal2006,Sminchisescu2007,Bo2009,Bo2010} or build a 3D model and then compute 3D pose by a generative model-image alignment process~\citep{Deutscher2000,Sidenbladh2000,Sminchisescu2001,Sminchisescu2003,Sminchisescu2005a,Sigal2004,Li2006,Andriluka2010,Gall2010}.
Recent methods tend to rely on the CNN architectures or successful 2D body joint detectors, both of which are usually discriminatively trained on a large amount of data.
We classify the recent methods into two classes (i.e., prediction-based and optimization-based methods) and review them in the following paragraphs.

The \emph{prediction-based approaches}~\citep{Ionescu2014a,Ionescu2014,Li2014,Tekin2016,Brau2016} directly estimate the 3D human pose from a 2D image.
In~\citep{Ionescu2014}, the authors present a large-scale structured prediction method that leverages the Fourier approximation of 2D histogram of oriented gradients (HOG) features for kernel dependency estimation.
In~\citep{Ionescu2014a}, the 3D human pose is regressed based on three stages: intermediate 2D body part labeling, second-order label-sensitive pooling, and 3D pose prediction by using pooled descriptors.
In~\citep{Li2014}, CNNs are proposed to solve a regression problem in which a mapping function from 2D image space to 3D pose space is learned based on two strategies: multi-task learning and pre-training approaches using 2D body part detectors.
In~\citep{Tekin2016}, motion information from consecutive frames of a video sequence is exploited to learn a regression function that directly predicts the 3D pose in a given frame of a sequence from a spatio-temporal volume centered on it.
In~\citep{Brau2016}, the authors propose to directly regress the 3D human pose and camera parameters based on a CNN, which is learned without the ground-truth 3D pose data by minimizing the loss containing the 2D joint annotations, 3D limb size constraints, and prior knowledge on 3D poses.
All these methods can produce the output 3D human pose with high efficiency because complicated optimization processes are not necessary.

In the \emph{optimization-based approaches}~\citep{Simo-Serra2012,Simo-Serra2013,Kostrikov2014,Li2015,Yasin2016,Zhou2016,Bogo2016}, an energy function is built using the prior term and the (intermediate) results of the 2D pose estimation methods, and the 3D human pose is obtained by minimizing the energy function.
In~\citep{Simo-Serra2012}, the authors propose a two-step approach in which 3D poses are probabilistically sampled based on the noisy results of the 2D body part detector and then pruned to generate an accurate 3D pose using both geometric and anthropomorphic constraints.
In~\citep{Simo-Serra2013}, 2D and 3D pose estimation problems are jointly solved by a Bayesian formulation combining a generative latent variable model that constrains the possible 3D poses and a HOG-based discriminative model that constrains the 2D body parts locations.
In~\citep{Kostrikov2014}, 2D human pose estimation is not utilized and body joints in 3D space are directly estimated by a 3D pictorial structure framework in which depth sweep regression forests are used to compute the likelihoods of 3D joint locations.
In~\citep{Li2015}, the authors present a unified framework for maximum-margin structured learning with deep neural network, where a score function taking both an image and a 3D pose as inputs is learned and minimized to generate the output 3D human pose.
In~\citep{Yasin2016}, to alleviate the burden of acquiring accurate pairs of a 2D image and a 3D pose, the authors propose a dual-source approach that can incorporate 2D and 3D information from two different training sources (i.e., images with annotated 2D pose and 3D motion capture data).
In~\citep{Zhou2016}, from an input monocular RGB image sequence, the 3D human poses for all frames are recovered based on an expectation-maximization (EM) algorithm that combines image-based 2D body joint detection, 3D geometric pose priors using sparse representation, and temporal model.
In~\citep{Bogo2016}, the authors propose a two-step approach in which a CNN-based 2D pose estimation method called DeepCut~\citep{Pishchulin2016} is first used to obtain 2D joints and a 3D generative body shape model called SMPL~\citep{Loper2015} is fitted to them, thereby resulting in a 3D mesh that captures both 3D human pose and shape.
The optimization-based approaches tend to produce more accurate results compared to the prediction-based approaches.
Our proposed method also belongs to the optimization-based approaches.
Unlike them, however, our approach does not rely on iterative optimization processes such as the EM algorithm and 3D shape fitting, thereby enabling accurate and fast 3D human pose estimation.

\section{Proposed CRF Model}
\label{Sec:ProposedModel}

In this section, we present the proposed CRF model for 3D human pose estimation from a single RGB image.
Let $I_{0}$ denote an input RGB image.
We assume that an additional square bounding box $B$ enclosing the subject is also given.
Two popular datasets such as Human3.6M~\citep{Ionescu2014} and HumanEva~\citep{Sigal2010} provide the bounding box information using background subtraction, and many recent 3D human pose estimation approaches~\citep{Tekin2016,Li2015,Zhou2016} adopt that assumption.
Our proposed CRF model defines the energy function over 2D human poses as follows:
\begin{equation}\label{Eq:Energy}
  E(\mathbf{x};I_{0},B)=U(\mathbf{x};I_{0},B)+V(\mathbf{x}),
\end{equation}
where $\mathbf{x}=(\mathbf{x}_1,\ldots,\mathbf{x}_M)$ denotes the 2D pixel coordinates of $M$ body joints within $I_{0}$.
The 3D human pose $\mathbf{X}=(\mathbf{X}_1,\ldots,\mathbf{X}_M)$ defined in the camera coordinate system is obtained as a byproduct after minimizing the energy function in Equation~(\ref{Eq:Energy}) to infer the MAP estimate of the 2D pose.
We refer to the proposed CRF model as the \emph{pose-lifting CRF (PLCRF)} because it can simultaneously estimate the optimal 2D pose and lift it to the 3D pose.
We describe the unary likelihood term $U(\mathbf{x};I_{0},B)$ and the high-order 2D-3D pose consistency term $V(\mathbf{x})$ in the next subsections.

\subsection{Unary Likelihood Term Using CNN}
\label{Sec:CNN}

In the proposed model, the unary term is defined as the negative likelihoods of $M$ joints occurring in the locations $\mathbf{x}_1,\ldots,\mathbf{x}_M$.
To define the likelihoods, we utilize the 2D heat maps that can be obtained by various heat map regression networks in many recent 2D human pose estimation methods~\citep{Tompson2014,Tompson2015,Pfister2015,Pishchulin2016,Insafutdinov2016}.
A $256\times256$ image $I$ is generated by cropping and resizing the bounding box region in $I_{0}$ and is then used as an input to the heat map regression network.
We interpret the heat maps as the probability distributions of the joint locations and define the unary term as follows:
\begin{equation}\label{Eq:Unary}
  U(\mathbf{x};I_{0},B)=-\sum_{i=1}^{M}h_{i}(\mathbf{p}_{i};I),
\end{equation}
where $h_{i}(\cdot;I)$ and $\mathbf{p}_{i}$ denote the 2D heat map of the joint $i$ computed from the input image $I$ and the 2D location of the joint $i$ within the heat map, respectively.
We set the resolution of each heat map to $32\times32$ for reducing the size of the heat map regression networks.
The $M$ heat maps $H=(h_1,\ldots,h_M)$ are computed simultaneously rather than separately.
Therefore, the heat map regression network should regress the $M\times32\times32$ heat map volume $H$ from the $3\times256\times256$ input image $I$.
We utilize two networks: SpatialNet~\citep{Pfister2015} and DeeperCut~\citep{Insafutdinov2016}.
Let $U_{\mathrm{s}}(\mathbf{x};I_{0},B)$ and $U_{\mathrm{d}}(\mathbf{x};I_{0},B)$ denote the unary terms obtained by SpatialNet and DeeperCut, respectively.
These two unary terms are used independently for the proposed CRF model.

\begin{figure}[t]
\centering
\includegraphics[width=1.0\linewidth]{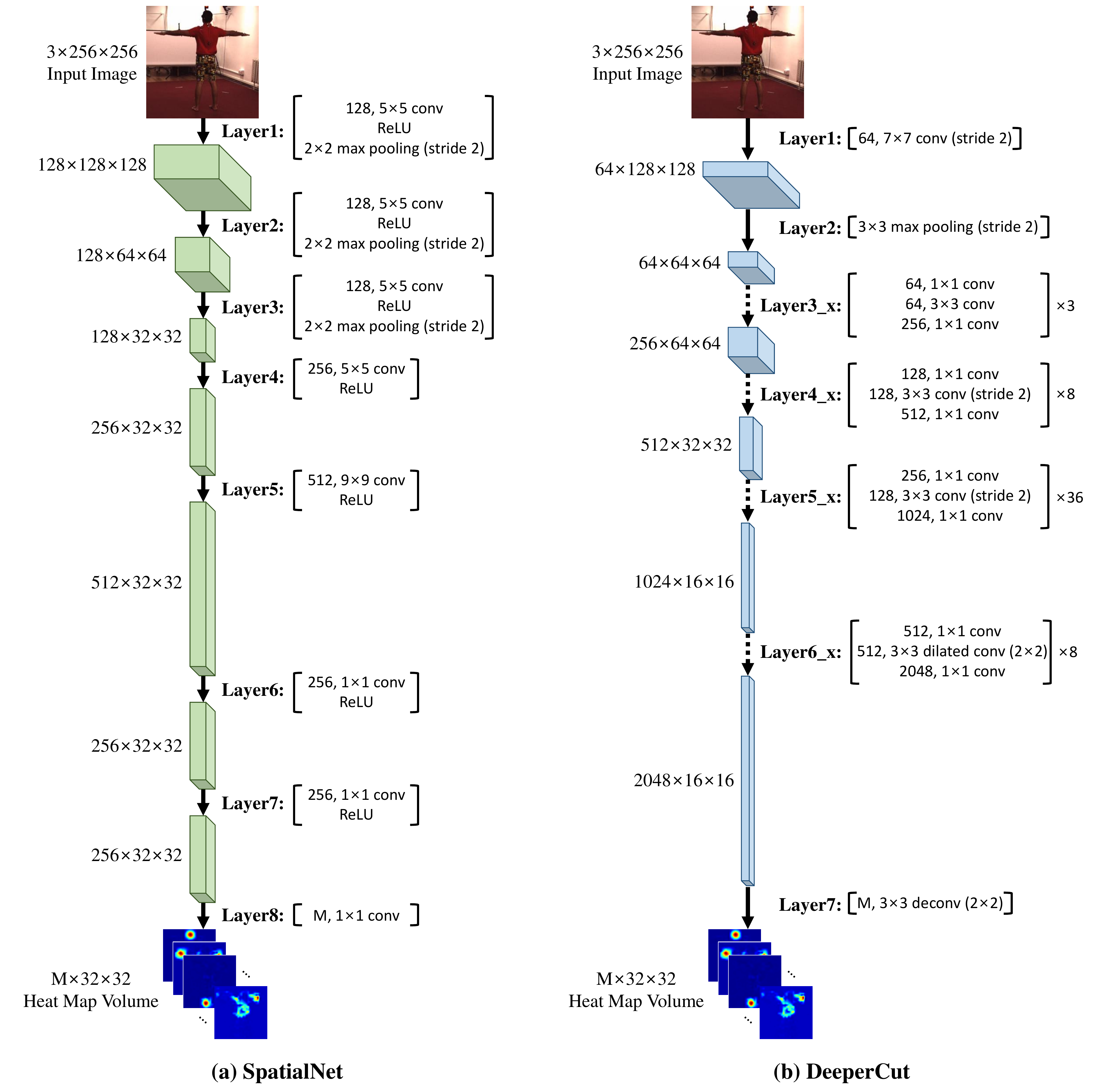}
\caption{Two heat map regression networks used in this study. Dotted arrows denote layers consisting of repetitions of same layer.}
\label{Fig:Networks}
\end{figure}

The architectures of the heat map regression networks are briefly described as follows:
\emph{SpatialNet} consists of seven convolutional layers with $5\times5$, $9\times9$, and $1\times1$ convolutional filters followed by a ReLU function and the last layer with an $M\times1\times1$ filter to produce the heat maps for all $M$ joints, as shown in Figure~\ref{Fig:Networks}(a).
A $2\times2$ max pooling operation is added after the first three convolutional layers to reduce the spatial resolution of the output heat map volume.
We do not use spatial fusion and temporal pooling as in~\citep{Zhou2016}.
\emph{DeeperCut} is obtained by modifying ResNet~\citep{He2016} with 152 layers, a state-of-the-art network for image classification.
Its structure is briefly illustrated in Figure~\ref{Fig:Networks}(b).
Note that batch normalization~\citep{Ioffe2015} and ReLU activation functions are applied sequentially after all convolutional operations.
The layers of Layer3, 4, 5, and 6 are building blocks for residual learning, and they contain shortcut connections in the form of identity mapping or simple linear mapping that are not shown in the figure.
The original ResNet contains five layers with stride 2, which results in a 32-pixel stride.
Therefore, we first change the stride of the $3\times3$ convolutional layer of the Layer6\_1 layer from 2.0 pixels to 1.0 pixel and then apply 2-dilated convolution~\citep{Yu2016} to maintain the size of its receptive field.
We also use the deconvolutional layer~\citep{Long2015} for 2x up-sampling as the Layer7 layer to make the total stride size of 8.0 pixels.
We do not combine the final output with the Layer4 output, nor do we use intermediate supervision, unlike the process for the original DeeperCut.
We empirically observed that this slight simplification does not significantly affect the performance of the proposed method.

We briefly describe how the heat map regression networks are learned.
Suppose that a training dataset consisting of pairs of a training image and its ground-truth 2D human pose is given.
To apply data normalization, we first compute the mean and standard deviation for each channel of training images.
Then, we subtract the mean values from the training images and divide the result by the standard deviation values.
The ground-truth heat map is modeled by an unnormalized Gaussian distribution with a standard deviation of 1.0 pixel centered on the ground-truth 2D joint location.
This allows small variations in estimating the 2D position of the joints, which in turn makes the heat map regressor easier to learn and robust to mistakes in annotating ground-truth 2D joints.
We also perform data augmentation by applying the following steps to each training image at each iteration for training.
First, a $248\times248$ image is randomly extracted from the training image $I$ and then resized to $256\times256$ pixels.
Then, each color channel is multiplied by a random value between 0.8 and 1.2 to jitter the contrast.
The networks are trained by minimizing the mean squared error (MSE) between the predicted and ground-truth heat maps.
For the SpatialNet, we use rmsprop~\citep{Tieleman2012} with a learning rate of $10^{-4}$ and a mini-batch size of 32.
We initialize the DeeperCut from the publicly available ResNet model pre-trained using the ImageNet dataset and fine-tune the DeeperCut using rmsprop with a learning rate of $10^{-5}$ and a mini-batch size of 16.
The number of epochs is set to 50 for both networks.

\begin{figure}[t]
\centering
\includegraphics[width=1.0\linewidth]{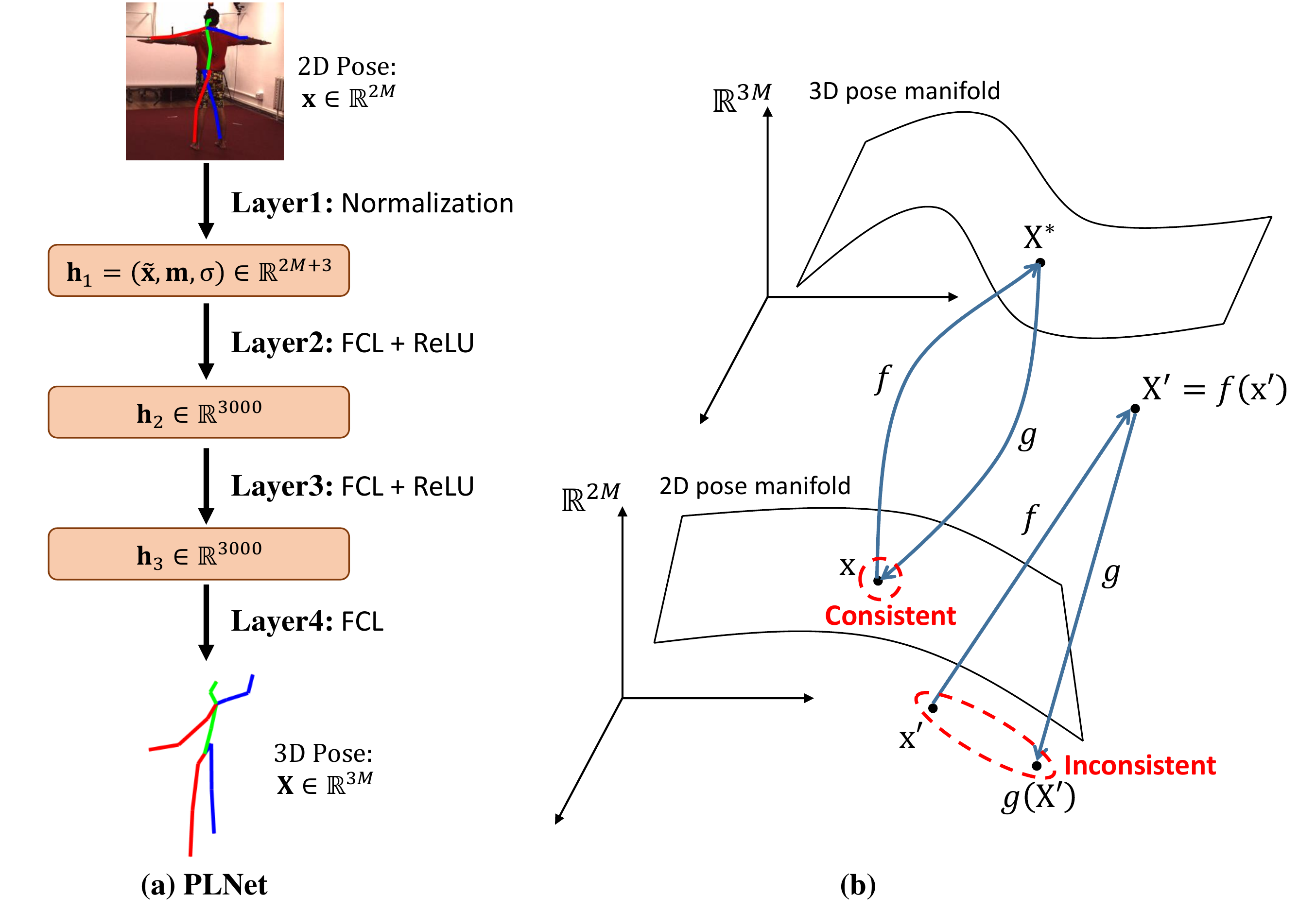}
\caption{The proposed 2D-to-3D pose lifting network is illustrated in (a) and the motivation of our 2D-3D pose consistency prior is described in (b). FCL denotes the fully-connected layer.}
\label{Fig:Intuition}
\end{figure}

\subsection{High-order 2D-3D Pose Consistency Term Using MLP}
\label{Sec:MLP}

In this subsection, we first propose a 2D-to-3D pose-lifting network mapping from 2D human pose $\mathbf{x}$ to 3D human pose $\mathbf{X}$ and present a 2D-3D pose consistency prior based on this.
We do not consider the absolute position of the output 3D pose, so we assume that $\mathbf{X}$ has zero mean.
As a preprocessing step, the input 2D pose $\mathbf{x}=(\mathbf{x}_1,\ldots,\mathbf{x}_M)$ is normalized to construct $\tilde{\mathbf{x}}=(\tilde{\mathbf{x}}_1,\ldots,\tilde{\mathbf{x}}_M)$ as follows:
\begin{equation}\label{Eq:Normalize}
  \tilde{\mathbf{x}}_{i}=\frac{\mathbf{x}_{i}-\mathbf{m}}{\sigma},
\end{equation}
where $\mathbf{m}=(\sum_{i=1}^{M}\mathbf{x}_{i})/M$ is the mean vector and $\sigma=\sqrt{(\sum_{i=1}^{M}\|\mathbf{x}_{i}-\mathbf{m}\|_{2}^{2})/M}$ is the standard deviation.
Unlike the orthographic camera projection case where $\mathbf{m}$ and $\sigma$ do not play a role in obtaining $\mathbf{X}$, they can provide additional information for the perspective camera projection to determine $\mathbf{X}$ correctly.
This is because $\mathbf{m}$ and $\sigma$ contain the information of the subject's 3D position that is used with the 3D pose $\mathbf{X}$ to generate the 2D pose $\tilde{\mathbf{x}}$.
Therefore, we feed $(\tilde{\mathbf{x}},\mathbf{m},\sigma)$ into the 2D-to-3D pose-lifting network as a ($2M+3$)-dimensional input vector.
To estimate $3M$-dimensional output vector $\mathbf{X}$, we simply use a three-layer MLP as illustrated in Figure~\ref{Fig:Intuition}(a).
In the proposed network, the number of hidden units for each layer is set to 3,000 and ReLU is adopted as an activation function.
We refer to this network as the \emph{pose-lifting network (PLNet)}.

We briefly describe how to learn the PLNet.
The ground-truth 3D poses for training are first normalized to have zero means.
For data augmentation, we add a Gaussian noise with zero mean and a standard deviation of 0.1 to the normalized training 2D pose $\tilde{\mathbf{x}}$ at each iteration of the training procedure.
This process allows the proposed network to be robust to the noisy estimation from conventional 2D human pose estimation methods.
We minimize the MSE between the predicted 3D pose and the ground-truth 3D pose using the stochastic gradient descent method in which the learning rate, momentum, and number of epochs are set to $10^{-3}$, 0.9, and 200, respectively.

Let $\Omega\subset\mathbb{R}^{2M}$ and $\Sigma\subset\mathbb{R}^{3M}$ be the normal 2D and 3D human pose spaces, respectively.
Multiple 3D human poses can be projected into a single 2D human pose due to the generic ambiguities of the 3D-to-2D projection process.
Therefore, a single 2D human pose cannot decide a 3D human pose deterministically and can correspond to many possible 3D human poses, in which some poses will be more probable than others.
And this results in a probability distribution function $p(\mathbf{X}|\mathbf{x})$ of the 3D human pose $\mathbf{X}\in\Sigma$ given a 2D human pose $\mathbf{x}\in\Omega$.
Let $f$ represent a function that maps from $\mathbf{x}$ to the point estimate $\mathbf{X}^{\ast}=\arg\max_{\mathbf{X}}p(\mathbf{X}|\mathbf{x})$.
We conjecture that the proposed PLNet can learn such a function $f$.
Suppose the function $g$ represents 3D-to-2D projection.
The composite function $g\circ{f}$ then represents the sequential process of 2D-to-3D lifting and 3D-to-2D projection.
A normal 2D human pose, $\mathbf{x}\in{\Omega}$, will be mapped back to $\mathbf{x}$ by such $g\circ{f}$.
However, the abnormal 2D human pose $\mathbf{x}\notin{\Omega}$ is less likely to be mapped to the original $\mathbf{x}$ by $g\circ{f}$, which is illustrated in Figure~\ref{Fig:Intuition}(b).
Therefore, we can estimate whether $\mathbf{x}$ is a normal 2D pose or not by comparing $\mathbf{x}$ with $(g\circ{f})(\mathbf{x})$.

The above idea of 2D-3D pose consistency is used to formulate the prior term $V(\mathbf{x})$ of the given 2D pose $\mathbf{x}$.
Under the perspective projection assumption, computing the re-projected 2D pose $\mathbf{y}=(g\circ{f})(\mathbf{x})$ requires the absolute 3D pose corresponding to $\mathbf{x}$.
However, since our PLNet is learned using the zero-mean 3D poses, its output $\mathbf{X}$ is not the absolute 3D pose.
Therefore, we estimate the average 3D position $\bar{\mathbf{M}}$ of the subject in the training data and use it to approximately compute the absolute pose $\bar{\mathbf{X}}=\mathbf{X}+\bar{\mathbf{M}}$.
Assuming that the camera calibration information is given, we can compute $\mathbf{y}=g_{\mathrm{pers}}(\bar{\mathbf{X}})$ in which $g_{\mathrm{pers}}$ involves the camera calibration parameters.
To address the uncertainty in $\bar{\mathbf{X}}$, we normalize $\mathbf{x}$ and $\mathbf{y}$ using Equation~(\ref{Eq:Normalize}) to be independent of 2D position and scale to produce $\tilde{\mathbf{x}}$ and $\tilde{\mathbf{y}}$, which are compared to define the prior term $V(\mathbf{x})$ in Equation~(\ref{Eq:Energy}) for 2D human pose $\mathbf{x}$ as follows:
\begin{equation}\label{Eq:PriorPers}
  V_{\mathrm{p}}(\mathbf{x})=\lambda\sum_{i=1}^{M}\|\tilde{\mathbf{x}}_{i}-\tilde{\mathbf{y}}_{i}\|_{2}^{2},
\end{equation}
where $\tilde{\mathbf{y}}$ is the result of applying the normalization process in Equation~(\ref{Eq:Normalize}) to $\mathbf{y}=g_{\mathrm{pers}}(f(\mathbf{x})+\bar{\mathbf{M}})$ and $\lambda$ denotes the parameter controlling the strength of the prior term.
Even if no camera calibration information is given, we can define the prior term by assuming orthographic projection $g_{\mathrm{ortho}}$ as follows:
\begin{equation}\label{Eq:PriorOrtho}
  V_{\mathrm{o}}(\mathbf{x})=\lambda\sum_{i=1}^{M}\|\tilde{\mathbf{x}}_{i}-\tilde{\mathbf{z}}_{i}\|_{2}^{2},
\end{equation}
where $\tilde{\mathbf{z}}$ is the result of applying the normalization process to $\mathbf{z}=g_{\mathrm{ortho}}(f(\mathbf{x}))$.
Note that $g_{\mathrm{ortho}}$ is a remarkably simple mapping that does not require any parameters.

\section{Inference Method}
\label{Sec:InferenceMethod}

This section describes how to obtain the MAP estimate of the proposed PLCRF model, that is, the optimal 2D human pose $\mathbf{x}^{\ast}$, by minimizing the energy function defined by Equation~(\ref{Eq:Energy}).
The optimal 3D human pose $\mathbf{X}^{\ast}$ can be obtained by feeding such $\mathbf{x}^{\ast}$ into our PLNet, which is actually performed in the process of minimizing the energy function.
However, the high-order prior term makes it difficult for the proposed energy function to be minimized accurately.
Therefore, we apply the \emph{N-best strategy}~\citep{Park2011} to obtain an approximate solution.

The N-best algorithm allows N-best candidates, i.e., $N$ lowest-energy solutions, to be efficiently generated for particular types of CRF models of which second-best solution can be easily determined, such as a pairwise CRF of the tree structure.
However, for computer vision problems such as 2D human pose estimation, second-best configurations are generally one-pixel shifted versions of the best.
Therefore, the authors of~\citep{Park2011} proposed a way to produce diverse solutions by adding constraints that N-best candidates should not overlap spatially.
In general, a better solution can be found by applying the high-order priors to these N-best candidates.

The proposed method also finds N-best candidates first from a simple model and produces an approximate solution of the original complex model in Equation~(\ref{Eq:Energy}) using them.
However, the proposed CRF model does not include a pairwise term.
Therefore, we apply the N-best algorithm to the decomposable data term for each joint.
We first find the N-best joint candidates for each joint and combine the candidates for all the joints to generate the N-best poses (i.e., sets of joints).
We describe each of these processes in the following subsections.

\subsection{N-Best Joint Candidate Generation Using Mean Shift}

For each joint $i$, we want to find N-best joints $\mathbf{q}_{i}^{(1)},\ldots,\mathbf{q}_{i}^{(N)}$ that maximize the $32\times32$ heat map $h_{i}(\mathbf{p};I)$.
We also want the joints not to overlap as pointed out in~\citep{Park2011}.
For that purpose, we propose to use the local maxima, i.e., modes, of the 2D heat map as such joints.
However, the resolution of the 2D heat map $h_{i}(\mathbf{p};I)$ is too low, so obtaining a smoothed distribution through kernel density estimation and using the modes defined on it are better options.

Let $\mathbf{p}_{j}$, $j=1,\ldots,n$ denote all pixel coordinates in the heat map of joint $i$.
We consider those points with a weight of $h_{i}(\mathbf{p}_{j};I)$ as data samples and find the unknown distribution that draws the weighted samples.
Its kernel density estimator can be written as
\begin{equation}\label{Eq:KDE}
  \hat{h}_{i}(\mathbf{p})\propto\sum_{j=1}^{n}h_{i}(\mathbf{p}_{j};I)\cdot{}k\left(\frac{\|\mathbf{p}-\mathbf{p}_{j}\|_{2}^{2}}{b^2}\right),
\end{equation}
where $\mathbf{p}$ is the 2D subpixel location within the $32\times32$ heat map, $k(\cdot)$ is the kernel function, and $b$ is the bandwidth.

The modes of distribution in Equation~(\ref{Eq:KDE}) can be obtained by using the well-known mode-seeking algorithm called \emph{mean shift}~\citep{Comaniciu2002}.
Under the assumption of a flat kernel, the weighted mean for the samples within the kernel window can be computed as
\begin{equation}\label{Eq:WeightedMean}
  m(\mathbf{p})=\frac{\sum_{\mathbf{p}_{j}\in\mathcal{N}(\mathbf{p})}h(\mathbf{p}_{j};I)\cdot\mathbf{p}_{j}}{\sum_{\mathbf{p}_{j}\in\mathcal{N}(\mathbf{p})}h(\mathbf{p}_{j};I)},
\end{equation}
where $\mathcal{N}(\mathbf{p})=\{\mathbf{p}_{j}:\|\mathbf{p}_{j}-\mathbf{p}\|_{2}<b\}$ is the neighborhood of $\mathbf{p}$.
The mean-shift algorithm starts with an initial estimate $\mathbf{q}$ and repeats the process of setting $\mathbf{q}\leftarrow{m(\mathbf{q})}$ until $m(\mathbf{q})$ converges.
We can obtain multiple modes by applying the aforementioned procedure to all pixels $\mathbf{p}_{j}$, $j=1,\ldots,n$, and merging the converged weighted means.
We can also obtain the smoothed heat map value $\hat{h}_{i}(\mathbf{q})$ corresponding to each mode $\mathbf{q}$ in that process.
By sorting these values, we can find N-best joints $\mathbf{q}_{i}^{(1)},\ldots,\mathbf{q}_{i}^{(N)}$ that satisfy $\hat{h}_{i}(\mathbf{q}_{i}^{(1)})\geq\ldots\geq\hat{h}_{i}(\mathbf{q}_{i}^{(N)})$.

\subsection{N-Best Pose Candidate Generation Using N-Best Algorithm}

Let $\mathcal{J}_{i}=\{\mathbf{q}_{i}^{(1)},\ldots,\mathbf{q}_{i}^{(N)}\}$ be the set of N-best joints for joint $i$, and let $\mathcal{P}=\mathcal{J}_{1}\times\cdots\times\mathcal{J}_{M}$ be the Cartesian product of such sets $\mathcal{J}_1,\ldots,\mathcal{J}_M$.
We can define the score function $S$ on $\mathcal{P}$, considering the smoothed heat map values, as follows:
\begin{equation}\label{Eq:Score}
  S(\mathbf{p})=\sum_{i=1}^{M}\hat{h}_{i}(\mathbf{p}_{i}),
\end{equation}
where $\mathbf{p}=(\mathbf{p}_{1},\ldots,\mathbf{p}_{M})\in{\mathcal{P}}$.
Now, our goal is to find $N$ highest-scoring poses $\mathbf{p}^{(1)},\ldots,\mathbf{p}^{(N)}$ from $\mathcal{P}$, the set of all possible poses.
The brute-force search is infeasible due to $|\mathcal{P}|=N^{M}$, so we apply the N-best algorithm~\citep{Yanover2004} to this problem.

\begin{algorithm}
\caption{Algorithm for generating N-best pose candidates}
\begin{algorithmic}
  \State {Initialize the best pose and its score:}
  \State {~~~~$\mathbf{p}^{(1)}\leftarrow{}(\mathbf{q}_{1}^{(1)},\ldots,\mathbf{q}_{M}^{(1)})$,~~$S^{(1)}\leftarrow{}\sum_{i=1}^{M}\hat{h}_{i}(\mathbf{q}_{i}^{(1)})$}
  \State {Initialize the set containing the best pose: $\mathcal{P}^{(1)}\leftarrow{}\mathcal{P}$}
  \For {$n=2,\ldots,N$}
    \State {1. Find the second-best pose $\dot{\mathbf{p}}^{(k)}$ of each set $\mathcal{P}^{(k)}$}
    \State {~~~~for $k=1,\ldots,n-1$:}
    \State {~~~~$\dot{\mathbf{p}}^{(k)}\leftarrow{}\mathrm{SecondBest}(\mathcal{P}^{(k)})$}
    \State {2. Find the second-best pose with the maximum}
    \State {~~~~score:}
    \State {~~~~$k'\leftarrow{}\arg\max_{k}S(\dot{\mathbf{p}}^{(k)})$}
    \State {3. Set the \emph{n}th-best pose and its score:}
    \State {~~~~$\mathbf{p}^{(n)}\leftarrow{}\dot{\mathbf{p}}^{(k')}$,~~$S^{(n)}\leftarrow{}S(\dot{\mathbf{p}}^{(k')})$}
    \State {4. Divide the set $\mathcal{P}^{(k')}$ into two disjoint sets $\mathcal{P}^{(n)}$}
    \State {~~~~and $\mathcal{P}^{(k')}$ in which $\mathbf{p}^{(n)}$ and $\mathbf{p}^{(k')}$ are the best}
    \State {~~~~poses, respectively}
  \EndFor
\end{algorithmic}
\label{Alg:NBest}
\end{algorithm}

The pseudocode for the N-best algorithm is briefly described in Algorithm~\ref{Alg:NBest}.
We provide the following additional explanation for steps 1 and 4.
Let us consider $\mathcal{P}$ instead of set $\mathcal{P}^{(k)}$ in step 1 without losing generality and investigate how to find the second-best pose in $\mathcal{P}$.
The best pose of set $\mathcal{P}$ is obviously $\mathbf{p}^{(1)}=(\mathbf{q}_{1}^{(1)},\ldots,\mathbf{q}_{M}^{(1)})$.
Candidates for the second-best pose can be obtained by selecting the second-best joint candidate for each joint.
In this case, the decrease in score for joint $i$ is $\hat{h}_{i}(\mathbf{q}_{i}^{(1)})-\hat{h}_{i}(\mathbf{q}_{i}^{(2)})$.
Thus, the second-best pose can be determined by searching the joint with the smallest decrease in score as $i'=\arg\min_{i}(\hat{h}_{i}(\mathbf{q}_{i}^{(1)})-\hat{h}_{i}(\mathbf{q}_{i}^{(2)}))$ and changing the position of such joint to $\mathbf{q}_{i'}^{(2)}$, which results in the second-best pose as follows:
\begin{equation}\label{Eq:SecondBest1}
  \mathrm{SecondBest}(\mathcal{P})=(\mathbf{q}_{1}^{(1)},\ldots,\mathbf{q}_{i'}^{(2)},\ldots,\mathbf{q}_{M}^{(1)}).
\end{equation}
Using the second-best pose, we can divide $\mathcal{P}$ into new $\mathcal{P}$ and $\mathcal{P}'$ in step 4 as follows:
\begin{equation}\label{Eq:SecondBest2}
  \left\{
    \begin{array}{lll}
      \mathcal{P}' & \leftarrow{} & \left\{\mathbf{p}:(\mathbf{p}\in\mathcal{P})\wedge(\mathbf{p}_{i'}=\mathbf{q}_{i'}^{(2)})\right\} \\
      \mathcal{P}  & \leftarrow{} & \mathcal{P}\setminus\mathcal{P}'
    \end{array}.
  \right.
\end{equation}
The output of Algorithm~\ref{Alg:NBest} consists of the N-best pose candidates $\mathbf{p}^{(1)},\ldots,\mathbf{p}^{(N)}$ and their corresponding scores $S^{(1)},\ldots,S^{(N)}$.

\subsection{Determining Optimal 2D and 3D Human Poses}

We can obtain the 2D pose candidates $\mathbf{x}^{(1)},\ldots,\mathbf{x}^{(N)}$ within the image $I_0$ from $\mathbf{p}^{(1)},\ldots,\mathbf{p}^{(N)}$.
The energy of each pose candidate $\mathbf{x}^{(k)}$ can be calculated as
\begin{equation}\label{Eq:NBestEnergy}
  E^{(k)}=E(\mathbf{x}^{(k)};I_{0},B)=-S^{(k)}+V(\mathbf{x}^{(k)}),
\end{equation}
where the second equality can be easily derived from Equations~(\ref{Eq:Unary}) and~(\ref{Eq:Score}).
Note that the 3D pose candidate $\mathbf{X}^{(k)}$ is obtained in the process of calculating $V(\mathbf{x}^{(k)})$.
Finally, the optimal 2D and 3D human poses $\mathbf{x}^{\ast}$ and $\mathbf{X}^{\ast}$ can be determined by finding the minimum energy as follows:
\begin{equation}\label{Eq:Inference}
  \left\{
    \begin{array}{lll}
      k^{\ast}          & = & \arg\min_{k}E^{(k)} \\
      \mathbf{x}^{\ast} & = & \mathbf{x}^{(k^{\ast})} \\
      \mathbf{X}^{\ast} & = & \mathbf{X}^{(k^{\ast})} \\
    \end{array}.
  \right.
\end{equation}

\section{Experimental Results}
\label{Sec:ExperimentalResults}

\subsection{Dataset}

We use two publicly available datasets to evaluate the proposed 3D human pose estimation method.
The first dataset is the recently released, large-scale dataset Human3.6M\footnote{http://vision.imar.ro/human3.6m}~\citep{Ionescu2014}.
The marker-based motion capture system is synchronized with four multi-view RGB cameras to generate approximately 3.6 million 3D poses and their corresponding 2D RGB images at 50 fps.
To create this dataset, 11 actors dressed in moderately realistic clothes perform various daily activities in an indoor environment, which includes 15 scenarios: \emph{directions, discussion, eating, greeting, phoning, taking photo, posing, purchases, sitting, sitting down, smoking, waiting, walk dog, walking, and walk together}.
The second dataset is the HumanEva-I dataset\footnote{http://humaneva.is.tue.mpg.de}~\citep{Sigal2010}, which uses a motion capture system and three RGB cameras.
For training and validation, approximately 13,600 frames of synchronized 3D pose and multi-view RGB images are generated at 60 fps.
Moreover, an additional training dataset consisting of only 3D pose data of about 37,000 frames acquired at 120 Hz is also available.
In this dataset, four subjects with natural clothing perform six predefined actions: \emph{walking, jogging, gesturing, throwing and catching a ball, boxing, and combo}.

\subsection{Evaluation Metric}

The proposed algorithm produces 2D and 3D human pose estimates $\mathbf{x}^{\ast}$ and $\mathbf{X}^{\ast}$ for a given input 2D image $I_{0}$.
Let us assume that the ground-truth 2D and 3D poses are $\mathbf{x}$ and $\mathbf{X}$.
To calculate the accuracy of the estimated 3D pose, we use the mean per-joint position error (MPJPE) applied in many studies~\citep{Ionescu2014,Li2014,Tekin2016,Li2015,Zhou2016}, which is as follows:
\begin{equation}
  J_{\mathrm{MPJPE}}=\frac{1}{M}\sum_{i=1}^{M}\|(\mathbf{X}_{i}-\mathbf{X}_{r})-(\mathbf{X}_{i}^{\ast}-\mathbf{X}_{r}^{\ast})\|_{2},
\end{equation}
where $r$ denotes the index of the root joint.
The aforementioned error is calculated for all samples in the test dataset, and their average is reported.
Note that this metric is computed in the relative coordinates with the root joint as the origin, not the absolute world coordinates.
We consider another error metric that is more relaxed than MPJPE and has been adopted in various studies~\citep{Simo-Serra2013,Kostrikov2014,Yasin2016,Bogo2016}.
We apply a similarity transformation to the estimated 3D pose to align it with the ground-truth 3D pose via the Procrustes analysis and define the following error:
\begin{equation}
  J_{\mathrm{Similarity}}=\frac{1}{M}\sum_{i=1}^{M}\|\mathbf{X}_{i}-\hat{\mathbf{X}}_{i}^{\ast}\|_{2},
\end{equation}
where $\hat{\mathbf{X}}^{\ast}$ is the estimated 3D pose after alignment.
In the case of the 2D pose estimate $\mathbf{x}^{\ast}$, the scale is different for each test image.
Therefore, to create a consistent evaluation metric, we adopt the following 2D error metric using the 2D pose estimate $\mathbf{p}^{\ast}$ defined on the normalized $256\times256$ image $I$:
\begin{equation}
  J_{\mathrm{2D}}=\frac{1}{M}\sum_{i=1}^{M}\|\mathbf{p}_{i}-\mathbf{p}_{i}^{\ast}\|_{2},
\end{equation}
where $\mathbf{p}$ is the ground-truth 2D pose within $I$.
Note that $J_{\mathrm{MPJPE}}$ and $J_{\mathrm{Similarity}}$ are in mm and $J_{\mathrm{2D}}$ is in pixel.

\subsection{Implementation Details}

According to existing studies~\citep{Ionescu2014,Tekin2016,Li2015,Zhou2016,Bogo2016}, the number of joints $M$ is set to 17 and 14 for the Human3.6M and HumanEva-I datasets, respectively.
The training process for the networks constituting the proposed PLCRF model and the related parameter settings are described in Section~\ref{Sec:ProposedModel}.
We use the open-source Torch7~\citep{Collobert2011} framework to implement all our networks.
Except the network parameters, the proposed method has three other controllable parameters: $\lambda$, $b$, and $N$.
The first parameter $\lambda$ that controls the strength of the prior term in Equations~(\ref{Eq:PriorPers}) and~(\ref{Eq:PriorOrtho}) is set to 1.0, which is satisfactory in all our experiments.
The second parameter is the bandwidth parameter $b$ for kernel density estimation and mean-shift algorithm in Equations~(\ref{Eq:KDE}) and~(\ref{Eq:WeightedMean}), which is set to 3.0.
The last parameter, $N$, indicates the number of cadidates in the N-best algorithm, which is set to 8 for the HumanEva-I dataset and 128 for the more challenging Human3.6M dataset.

\subsection{Performance Analysis for 2D-to-3D Pose Lifting}

\tabcolsep=4pt
\begin{table}[t]
\caption{3D errors (mm) of 2D-to-3D pose lifting methods.}
\begin{center}
\begin{tabular}{lcc}
\hline
Method & $J_{\mathrm{MPJPE}}$ & $J_{\mathrm{Similarity}}$ \\
\hline
Ramakrishna et al. \citep{Ramakrishna2012} & - & 89.50 \\
Dai et al. \citep{Dai2014} & - & 72.98 \\
Zhou et al. \citep{Zhou2015} & - & 50.04 \\
Zhou et al. \citep{Zhou2016} & - & 49.64 \\
\hline
PLNet with input $\mathbf{x}$ & 47.52 & 36.49 \\
PLNet with input $\tilde{\mathbf{x}}$ & 51.07 & 34.84 \\
PLNet with input $(\tilde{\mathbf{x}},\mathbf{m},\sigma)$ & \textBF{45.19} & \textBF{33.39} \\
\hline
\end{tabular}
\end{center}
\label{Table:2Dto3DResults}
\end{table}

This subsection presents the evaluation results of the proposed 2D-to-3D pose-lifting method.
To do so, we train the PLNet proposed in Section~\ref{Sec:MLP} using data from the five subjects (i.e., S1, S5, S6, S7, and S8) of the Human3.6M dataset.
Then, according to the protocol of~\citep{Zhou2016}, the frames within 30 seconds belonging to the sequences of S9 and S11 from the first camera are used for the evaluation.
Note that one general dataset is constructed with no distinction of action type, and the original frame rate (i.e., 50 fps) is downsampled to 10 fps for the evaluation.

First, we investigate the effect of the proposed data normalization process in Equation~(\ref{Eq:Normalize}).
To perform this task, we separately learn two PLNets that accept the raw 2D pose $\mathbf{x}$ and the normalized 2D pose $\tilde{\mathbf{x}}$ as inputs, respectively.
Table~\ref{Table:2Dto3DResults} shows the results of such networks.
Interestingly, the normalized pose $\tilde{\mathbf{x}}$ produces better $J_{\mathrm{Similarity}}$ results than the raw pose $\mathbf{x}$ despite the loss of mean 2D location and scale information, which shows that data normalization is effective for rotation and scale invariant 3D reconstruction.
However, the best performance can be achieved by adding the mean 2D location $\mathbf{m}$ and scale $\sigma$, which are essential pieces of information for perspective reconstruction, to the input data of our network, which justifies the proposed data normalization process.

Next, we perform quantitative comparisons with other 2D-to-3D pose-lifting methods, which are illustrated in Table~\ref{Table:2Dto3DResults}.
Baseline methods include the single-view pose reconstruction method~\citep{Ramakrishna2012} based on the projected matching pursuit algorithm and the state-of-the-art non-rigid structure from motion algorithm~\citep{Dai2014} that requires video as its input.
The recently proposed convex relaxation-based method~\citep{Zhou2015} achieves impressive 3D human pose estimation performance, and the method is combined with the video input in~\citep{Zhou2016} for improved results.
In this evaluation, we consider the general action rather than the specific types of actions that can provide additional information, thereby further increasing the ambiguity of 3D pose estimation.
Moreover, the proposed method does not utilize multiple frames or entire videos that can be helpful in reducing the estimation ambiguity.
Surprisingly, however, the proposed discriminative method based on a simple MLP significantly outperforms its competitors.

\begin{figure}[t]
\centering
\includegraphics[width=0.9\linewidth]{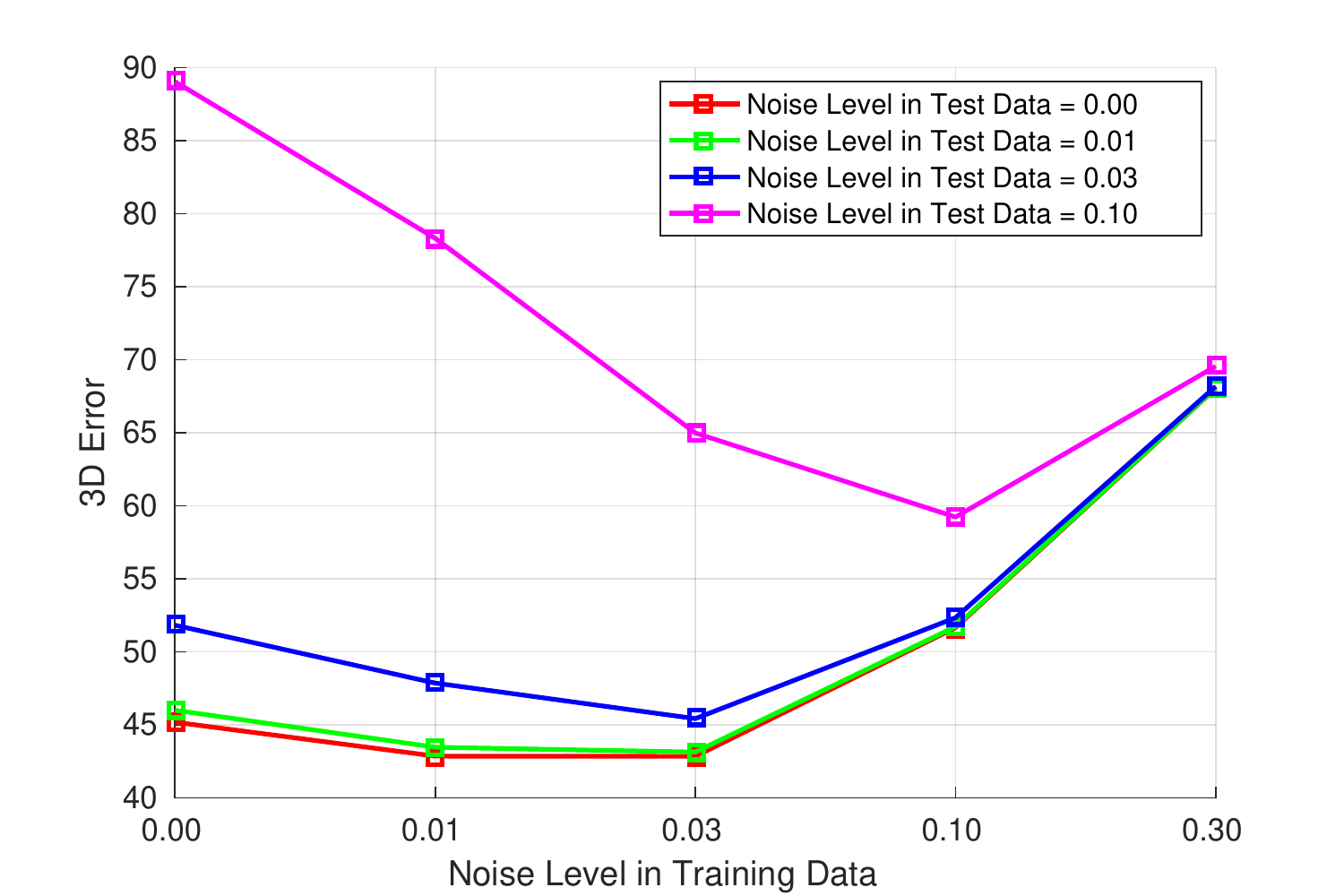}
\caption{Data augmentation by adding noise to training data helps in making the proposed method robust to noise in test data. The noise level means the standard deviation of the added zero-mean Gaussian noise and $J_{\mathrm{MPJPE}}$ is used to measure the 3D error in mm.}
\label{Fig:AnalysisNoise}
\end{figure}

We investigate the robustness of our method to the noise.
We add zero-mean Gaussian noise with various levels of standard deviations to the test data and apply the proposed method to such noisy datasets.
Figure~\ref{Fig:AnalysisNoise} illustrates the results of applying data augmentation (i.e., adding noise to training data) to the proposed method for robustness to test noise.
As expected, for the test data with a high noise level, the noise level of the training data should be also high.
Even for noise-free test data, training with little noise contributes to pose-lifting performance.

\subsection{Performance Analysis for Single Image 3D Pose Estimation}

\tabcolsep=1.5pt
\begin{table*}[t]
\small
\caption{3D errors (mm) of 3D human pose estimation methods for the Human3.6M dataset. ``NMS'' means that the non-maximum suppression is adopted for generating N-best pose candidates instead of the mean-shift algorithm.}
\begin{center}
\begin{tabular}{lcccccccc}
\hline
Method & Directions & Discussion & Eating & Greeting & Phoning & Photo & Posing & Purchases \\
\hline
Ionescu et al. \citep{Ionescu2014} & 132.71 & 183.55 & 132.37 & 164.39 & 162.12 & 205.94 & 150.61 & 171.31 \\
Li et al. \citep{Li2015} & - & 136.88 & 96.94 & 124.74 & - & 168.68 & - & - \\
Tekin et al. \citep{Tekin2016} & 102.39 & 158.52 & 87.95 & 126.83 & 118.37 & 185.02 & 114.69 & 107.61 \\
Zhou et al. \citep{Zhou2016} & 87.36 & 109.31 & 87.05 & 103.16 & 116.18 & 143.32 & 106.88 & 99.78 \\
PLCRF ($U_{\mathrm{s}}$) & 80.06 & 130.33 & 77.38 & 102.79 & 119.13 & 161.48 & 101.31 & 102.83 \\
PLCRF ($U_{\mathrm{s}}$+$V_{\mathrm{p}}$, NMS) & 77.87 & 120.85 & 75.91 & 100.41 & 114.62 & 153.91 & 97.20 & 98.92 \\
PLCRF ($U_{\mathrm{s}}$+$V_{\mathrm{p}}$) & 71.71 & 114.89 & 69.35 & 91.63 & 109.22 & 147.88 & 90.34 & 91.50 \\
PLCRF ($U_{\mathrm{s}}$+$V_{\mathrm{o}}$) & 71.80 & 114.86 & 69.51 & 91.88 & 109.42 & 148.35 & 90.49 & 92.02 \\
PLCRF ($U_{\mathrm{d}}$) & 65.77 & 85.36 & 62.52 & 79.16 & 84.35 & 113.26 & 76.08 & 68.46 \\
PLCRF ($U_{\mathrm{d}}$+$V_{\mathrm{p}}$) & \textBF{62.81} & \textBF{81.12} & \textBF{60.01} & 76.32 & \textBF{82.02} & \textBF{107.98} & \textBF{72.50} & \textBF{65.18} \\
PLCRF ($U_{\mathrm{d}}$+$V_{\mathrm{o}}$) & 62.82 & 81.17 & 60.03 & \textBF{76.29} & 82.71 & 108.18 & 72.58& 65.21 \\
\hline
       &         & Sitting &         &         & Walking &         & Walking  & \\[-3pt]
Method & Sitting & Down    & Smoking & Waiting & Dog     & Walking & Together & Average \\
\hline
Ionescu et al. \citep{Ionescu2014} & 151.57 & 243.03 & 162.14 & 170.69 & 177.13 & 96.60 & 127.88 & 162.14 \\
Li et al. \citep{Li2015} & - & - & - & - & 132.17 & 69.97 & - & - \\
Tekin et al. \citep{Tekin2016} & 136.15 & 205.65 & 118.21 & 146.66 & 128.11 & 65.86 & 77.21 & 125.28 \\
Zhou et al. \citep{Zhou2016} & 124.52 & 199.23 & 107.42 & 118.09 & 114.23 & 79.39 & 97.70  & 113.01 \\
PLCRF ($U_{\mathrm{s}}$) & 131.34 & 209.07 & 106.90 & 131.92 & 134.55 & 68.08 & 100.71 & 117.19 \\
PLCRF ($U_{\mathrm{s}}$+$V_{\mathrm{p}}$, NMS) & 124.78 & 193.28 & 102.18 & 127.56 & 125.28 & 70.17 & 100.19 & 112.21 \\
PLCRF ($U_{\mathrm{s}}$+$V_{\mathrm{p}}$) & 116.13 & 184.15 & 94.33 & 117.61 & 120.00 & 60.41 & 85.94 & 104.34 \\
PLCRF ($U_{\mathrm{s}}$+$V_{\mathrm{o}}$) & 117.09 & 183.16 & 94.57 & 117.61 & 120.35 & 60.42 & 85.95 & 104.50 \\
PLCRF ($U_{\mathrm{d}}$) & 94.12 & 112.55 & 72.21 & 97.26 & 82.31 & 56.99 & 60.77 & 80.75 \\
PLCRF ($U_{\mathrm{d}}$+$V_{\mathrm{p}}$) & \textBF{90.94} & \textBF{104.13} & \textBF{69.49} & 93.06 & 78.13 & 54.45 & \textBF{58.18} & \textBF{77.09} \\
PLCRF ($U_{\mathrm{d}}$+$V_{\mathrm{o}}$) & \textBF{90.94} & 105.69 & 69.51 & \textBF{92.93} & \textBF{78.08} & \textBF{54.43} & 58.24 & 77.25 \\
\hline
\end{tabular}
\end{center}
\label{Table:SingleImageResults}
\end{table*}

\begin{figure}[t]
\centering
\includegraphics[width=1.0\linewidth]{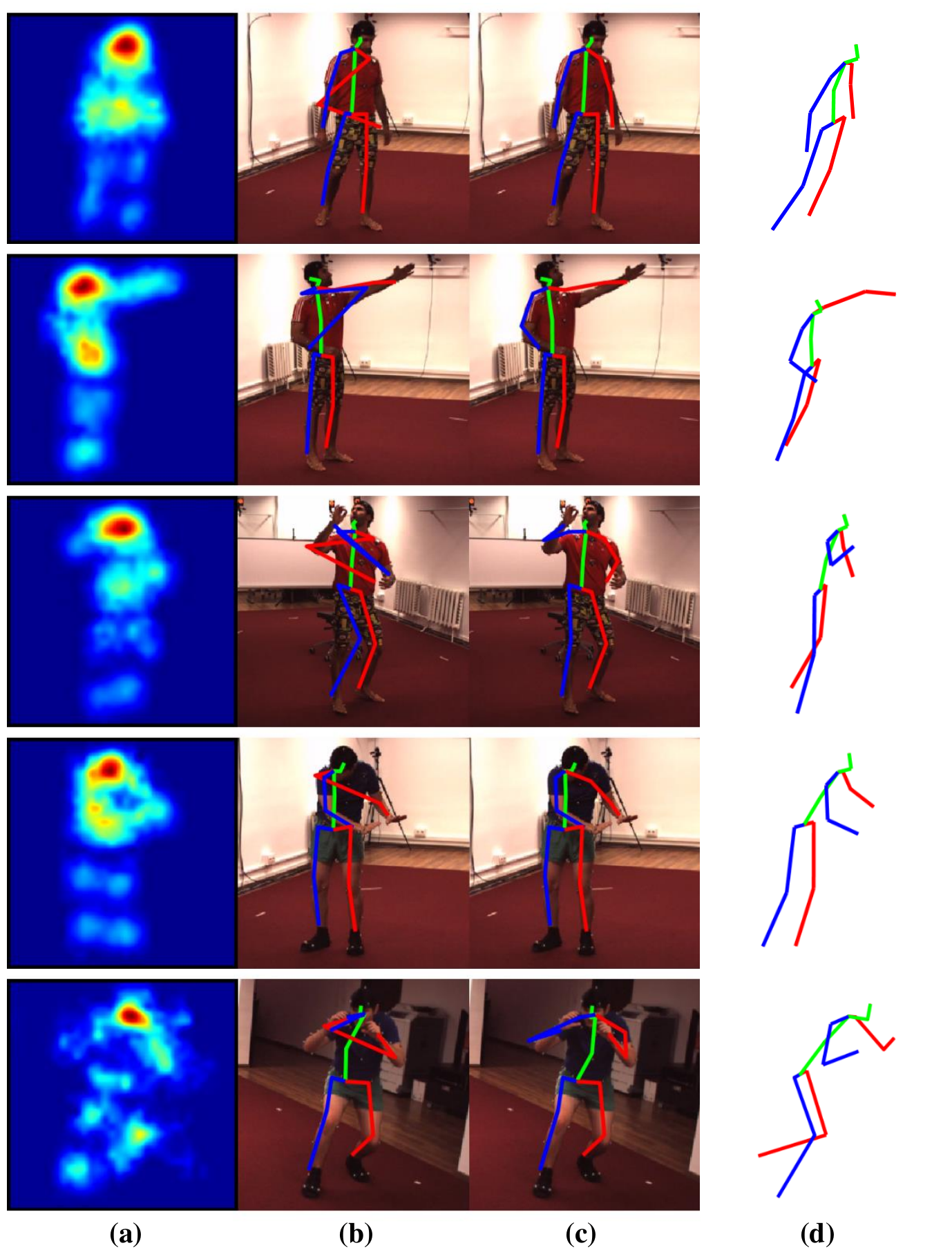}
\caption{Heat maps for all joints are combined and visualized in (a). The 2D pose consisting of greedily selected joints from the heat maps is shown in (b). The proposed method outputs the improved 2D pose estimate in (c) and its corresponding 3D pose in (d).}
\label{Fig:Images1}
\end{figure}

\begin{figure}[t]
\centering
\includegraphics[width=1.0\linewidth]{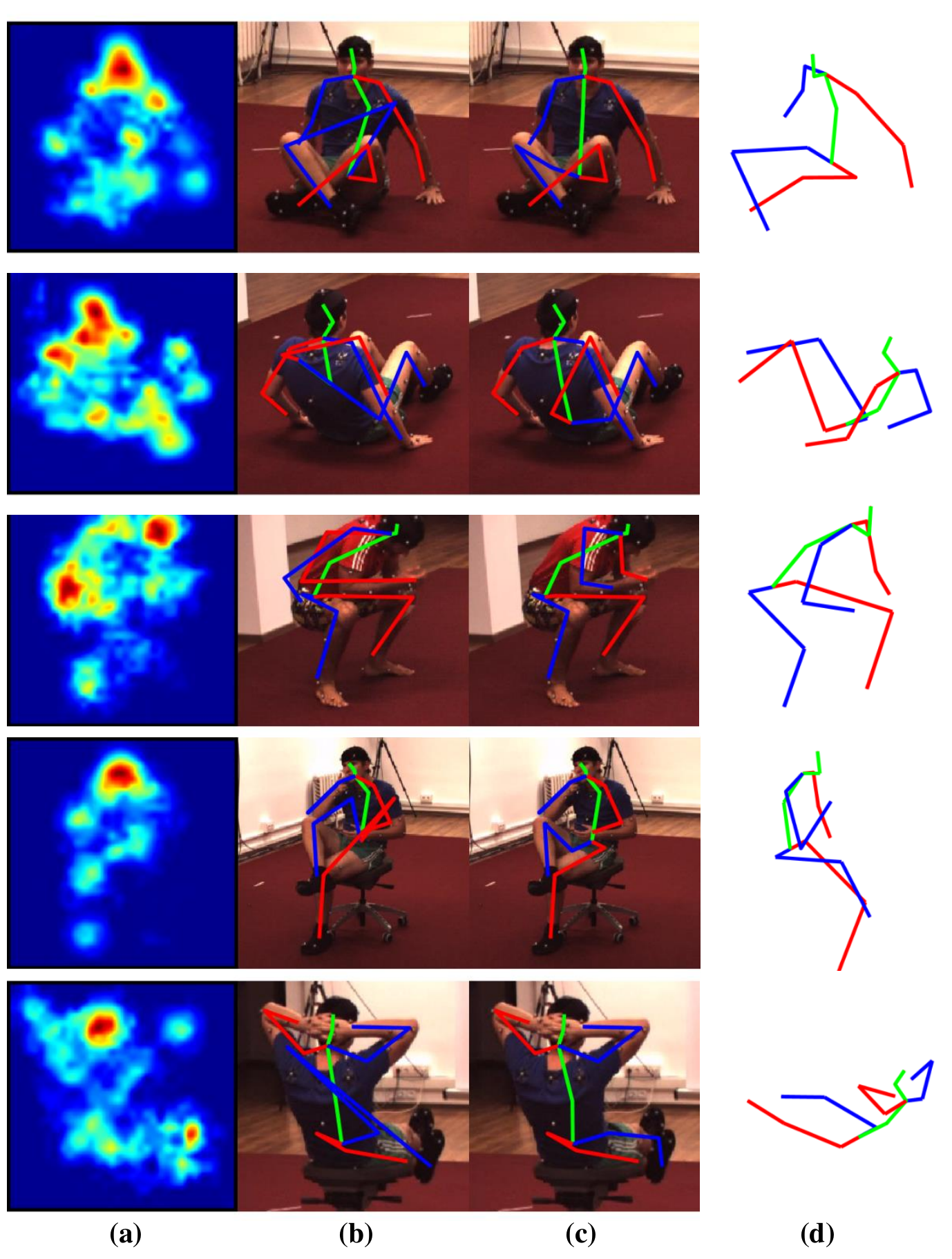}
\caption{Heat maps for all joints are combined and visualized in (a). The 2D pose consisting of greedily selected joints from the heat maps is shown in (b). The proposed method outputs the improved 2D pose estimate in (c) and its corresponding 3D pose in (d).}
\label{Fig:Images2}
\end{figure}

This subsection presents the evaluation results of the proposed PLCRF model for 3D human pose estimation from a single image.
To perform this task, according to the protocol of~\citep{Zhou2016}, we use the sequences of S9 and S11 from all cameras belonging to the Human3.6M dataset for evaluation.
Note that this evaluation is performed separately for each of the 15 action classes and all their MPJPE values are reported.
The results are illustrated in Table~\ref{Table:SingleImageResults}.
We first compute the best 2D pose by using the unary term $U_{\mathrm{s}}$ only and then estimate the 3D pose by applying the proposed PLNet, in which the proposed 2D-3D pose consistency term is not considered.
Next, we test the non-maximum suppression (NMS), which is often used in conventional human pose estimation methods, instead of the mean shift for generating N-best joint candidates.
To perform this task, we first resize each $32\times32$ heat map to $256\times256$ pixels and find all local maximum pixels, which are pruned to the N-best joint candidates by the NMS method.
We then use the proposed N-best algorithm to obtain the N-best pose candidates and apply the CRF energy including the proposed prior term, which produces more improved results (i.e., average error of 112.21 mm) than the unary term alone (117.19 mm).
We can see that the proposed N-best algorithm based on the mean shift can be used to further improve the performance (104.34 mm).
As expected, the mean shift seems to be more effective for joint detection than the NMS.
The 3D error is significantly reduced using the unary term $U_{\mathrm{d}}$ obtained by the 152-layer DeeperCut (77.09 mm), which shows the importance of the unary likelihood term.
Finally, we test the orthographic projection in Equation~(\ref{Eq:PriorOrtho}) rather than the perspective projection in Equation~(\ref{Eq:PriorPers}) to define the proposed prior term.
Obviously, the prior term $V_\mathrm{p}$ by the perspective projection achieves better performance.
However, the performance difference between them is very small (0.16 mm), which indicates the usefulness of the orthographic projection-based prior term $V_\mathrm{o}$ because it does not require the additional information of camera calibration.
Note that the proposed method also improves the 2D human pose estimation performance, which is illustrated in Table~\ref{Table:SingleImageResults2D}.
Figures~\ref{Fig:Images1} and~\ref{Fig:Images2} visualize the improved 2D poses and their corresponding 3D poses through the proposed method.

\tabcolsep=4pt
\begin{table}[t]
\caption{2D errors (pixel) of the proposed method for the Human3.6M dataset.}
\begin{center}
\begin{tabular}{lc}
\hline
Method & $J_{\mathrm{2D}}$ \\
\hline
PLCRF ($U_{\mathrm{s}}$) & 14.54 \\
PLCRF ($U_{\mathrm{s}}$+$V_{\mathrm{p}}$, NMS) & 14.07 \\
PLCRF ($U_{\mathrm{s}}$+$V_{\mathrm{p}}$) & 12.83 \\
PLCRF ($U_{\mathrm{s}}$+$V_{\mathrm{o}}$) & 12.85 \\
PLCRF ($U_{\mathrm{d}}$) & 9.84 \\
PLCRF ($U_{\mathrm{d}}$+$V_{\mathrm{p}}$) & \textBF{9.30} \\
PLCRF ($U_{\mathrm{d}}$+$V_{\mathrm{o}}$) & 9.32 \\
\hline
\end{tabular}
\end{center}
\label{Table:SingleImageResults2D}
\end{table}

We perform quantitative comparisons with other 3D human pose estimation methods.
Their results for the Human3.6M dataset are presented in Table~\ref{Table:SingleImageResults}.
The method in~\citep{Zhou2016} shows good performance (113.01 mm) by optimizing the entire video based on the EM algorithm, which depends on the heat map regression network, specifically the SpatialNet.
Thus, we consider the proposed method using the unary term $U_{\mathrm{s}}$ that is trained in the same way as in~\citep{Zhou2016} for a fair comparison.
The experimental results show that the proposed method performs more effectively (104.34 mm).
In the case of the more powerful unary term $U_{\mathrm{d}}$, the proposed method achieves a significantly better performance than all the other methods.
Note that our approach is based on a single image unlike the methods in~\citep{Tekin2016} and~\citep{Zhou2016} that utilize information from multiple frames.

\tabcolsep=2pt
\begin{table}[t]
\scriptsize
\caption{3D errors (mm) of 3D human pose estimation methods for the HumanEva-I dataset. ``MPII'' and ``HumanEva'' means that the MPII dataset and the HumanEva-I dataset are used to train our unary term $U_{\mathrm{d}}$, respectively. As the prior term, $V_{\mathrm{o}}$ is utilized.}
\begin{center}
\begin{tabular}{lccccccc}
\hline
\multicolumn{1}{l}{} & \multicolumn{3}{c}{Walking} & \multicolumn{3}{c}{Boxing} & \multicolumn{1}{c}{Average} \\
Method & S1 & S2 & S3 & S1 & S2 & S3 & \\
\hline
Akhter \& Black \citep{Akhter2015} & 186.1 & 197.8 & 209.4 & 165.5 & 196.5 & 208.4 & 194.4 \\
Ramakrishna et al. \citep{Ramakrishna2012} & 161.8 & 182.0 & 188.6 & 151.0 & 170.4 & 158.3 & 168.4 \\
Zhou et al. \citep{Zhou2015} & 100.0 & 98.9 & 123.1 & 112.5 & 118.6 & 110.0 & 110.0 \\
Bogo et al. \citep{Bogo2016} & 73.3 & 59.0 & 99.4 & 82.1 & 79.2 & 87.2 & 79.9 \\
PLCRF (MPII) & 39.4 & 37.6 & 74.7 & 45.8 & 56.4 & 60.5 & 52.5 \\
PLCRF (HumanEva) & \textBF{24.1} & \textBF{21.9} & \textBF{60.4} & \textBF{29.3} & \textBF{40.9} & \textBF{37.0} & \textBF{35.1} \\
\hline
\end{tabular}
\end{center}
\label{Table:HumanEvaResults}
\end{table}

We investigate the results of applying our algorithm to the HumanEva-I dataset and compare it with other methods.
According to the protocol in~\citep{Bogo2016}, the \emph{walking} and \emph{boxing} sequences of S1, S2, and S3 are used for evaluation.
The proposed method is first learned from the training data and then applied to the validation data to produce the evaluation results.
In~\citep{Bogo2016}, the authors compare several 3D human pose estimation methods using the 2D pose detector learned from another dataset (i.e., MPII human pose dataset~\citep{Andriluka2014}).
We also use the publicly available DeeperCut model\footnote{https://github.com/eldar/deepcut-cnn} that is learned from the MPII dataset as the heat map regression network for our method.
As the prior term, $V_{\mathrm{o}}$ is adopted because of its simplicity.
Table~\ref{Table:HumanEvaResults} provides the evaluation results up to the similarity transform (i.e., $J_{\mathrm{Similarity}}$).
The proposed method significantly outperforms other methods, which is a convincing result because, unlike other methods that utilize the resulting (hard) joints from the 2D pose detector, the proposed method relies on the (soft) joints in the form of the heat map.
Our method provides a mechanism for correcting erroneous high likelihood joints through the CRF optimization considering the 2D-3D pose consistency.
Note that using the unary term $U_{\mathrm{d}}$, which is learned from the HumanEva-I dataset, allows the proposed method to achieve the best performance.

\subsection{Computational Complexity}

All our experiments were performed on a desktop with Intel i7 3.5 GHz CPU, 128 GB RAM, and a Titan X GPU.
Learning the networks using the training sequences of the \emph{directions} class (6,387 frames) in the Human3.6M dataset took 14.6 hours for a SpatialNet, 39.5 hours for a DeeperCut, and 156 seconds for a PLNet.
Let us investigate the process of estimating 3D human pose from a single RGB image using the learned PLCRF model.
The running times for the three steps constituting the estimation process are as follows.
First, the running time for heat map regression is 21 milliseconds for SpatialNet and 25 milliseconds for DeeperCut.
Second, N-best candidate generation took 15 milliseconds.
Third, 2D-to-3D pose lifting and best-pose prediction took 6 milliseconds.
Therefore, the running time for estimating the 3D pose from a single image is 42-46 milliseconds, which shows the efficiency of the proposed method.

\section{Conclusion}
\label{Sec:Conclusion}

This study addressed the problem of 3D human pose estimation from a single RGB image.
We have shown that the simple MLP-based discriminative network (i.e., PLNet) is surprisingly effective in lifting 2D pose to 3D pose.
Moreover, by checking the consistency between the estimated 3D pose and the original 2D pose, the improbable solutions have been successfully suppressed.
We have also shown that the proposed high-order CRF model (i.e., PLCRF) can be efficiently optimized based on the N-best strategy.
The proposed method performs effectively for datasets acquired in a controlled laboratory environment, but whether it will work well for highly variable data in the wild is unclear.
Therefore, we aim to test the proposed approach in a more general environment to identify and solve related problems in the future.
We also aim to address the 3D pose estimation of multiple people and solve the problem of occlusions and ambiguities.

\section*{References}
\bibliographystyle{elsarticle-num}
\bibliography{mybib_20171108}

\end{document}